\documentclass[a4paper,conference]{IEEEtran}

\ifCLASSINFOpdf
  \usepackage[pdftex]{graphicx}
  \DeclareGraphicsExtensions{.pdf,.jpg,.png}
\else
\fi

\usepackage[caption=false,font=footnotesize]{subfig}

\usepackage{booktabs}

\def\MYTITLE{Stereo Co-capture System for Recording and Tracking Fish with Frame- and Event Cameras}

\usepackage[pagebackref=false,breaklinks=true,colorlinks,bookmarks=false]{hyperref}
\hypersetup{
  pdftitle={\MYTITLE},
  pdfsubject={Computer Vision, Biology},
  pdfauthor={Friedhelm Hamann, Guillermo Gallego},
  pdfkeywords={Event Cameras, Animal tracking, Asynchronous Sensor, High Temporal Resolution}
}

\usepackage[capitalize]{cleveref}
\crefname{section}{Sec.}{Secs.}
\Crefname{section}{Section}{Sections}
\Crefname{table}{Table}{Tables}
\crefname{table}{Tab.}{Tabs.}

\usepackage{tikz}
\usetikzlibrary{arrows, arrows.meta}
\usepackage{caption}
\usepackage[separate-uncertainty=true]{siunitx}

\definecolor{light-gray}{gray}{0.7}
\newcommand\gframe[1]{{\color{light-gray}\frame{#1}}}

\definecolor{somegray}{gray}{0.6}
\newcommand{\darkgrayed}[1]{\textcolor{somegray}{#1}}
\makeatletter
\newcommand*\titleheader[1]{\gdef\@titleheader{#1}}
\AtBeginDocument{%
  \let\st@red@title\@title
  \def\@title{%
    \vskip-1.45em
    \bgroup\normalfont\large\centering\@titleheader\par\egroup
    \vskip0.2em\st@red@title}
}
\makeatother
\titleheader{\darkgrayed{This paper has been accepted for publication at the 26th International Conference on\\ Pattern Recognition (ICPR) VAIB Workshop, Montreal, Canada, 2022.
\copyright IEEE}}

\title{\MYTITLE}

\begin{document}

\author{
\IEEEauthorblockN{Friedhelm Hamann and Guillermo Gallego}
\IEEEauthorblockA{Technische Universit\"at Berlin, Einstein Center Digital Future and SCIoI Excellence Cluster, Berlin, Germany}}

\twocolumn[{%
\renewcommand\twocolumn[1][]{#1}%
\maketitle
\centering
\vspace{-0.1cm}
\def\eyefigheight{4.1cm}
\gframe{\includegraphics[height=\eyefigheight]{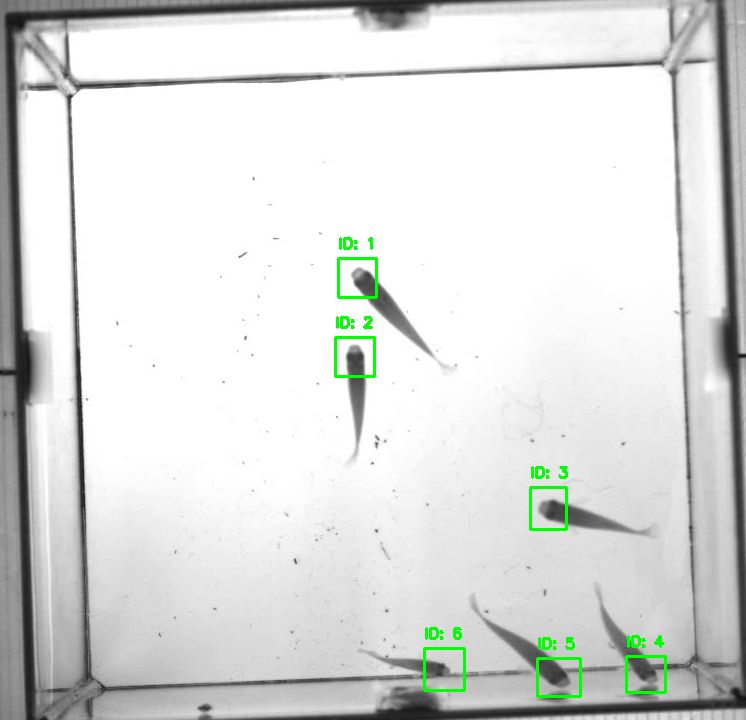}}\quad
\gframe{\includegraphics[height=\eyefigheight]{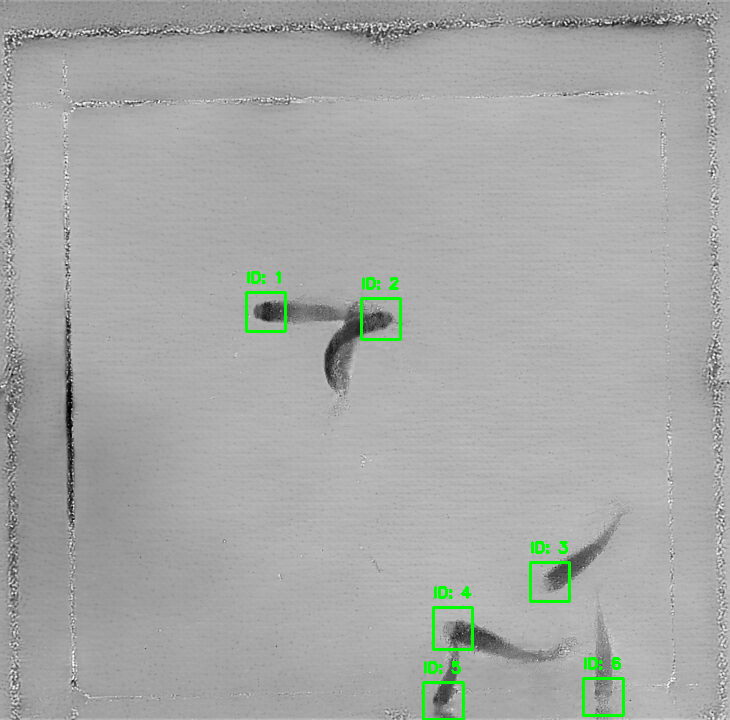}}\quad
\gframe{\includegraphics[height=\eyefigheight]{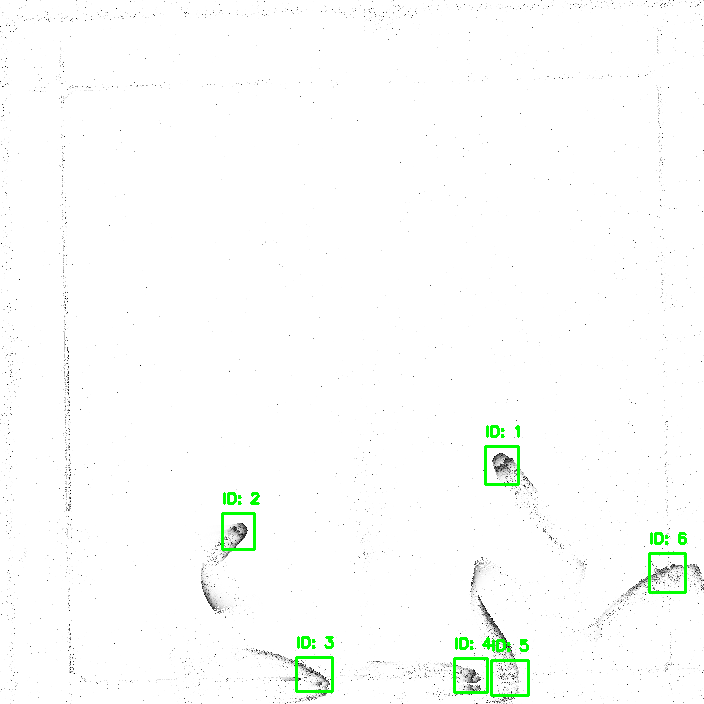}}
\vspace{0.3cm}
\captionof{figure}{\label{fig:eyecatcher}Visualization of the tracking output on different sensor data. 
From left to right: tracking on grayscale frames of a conventional camera, on images reconstructed from event data using E2VID~\cite{Rebecq19pami} and on time maps obtained from event data.
\vspace{0.4cm}
}
}]

\begin{abstract}

This work introduces a co-capture system for multi-animal visual data acquisition using conventional cameras and event cameras. 
Event cameras offer multiple advantages over frame-based cameras, such as a high temporal resolution and temporal redundancy suppression, which enable us to efficiently capture the fast and erratic movements of fish. 
We furthermore present an event-based multi-animal tracking algorithm, which proves the feasibility of the approach and sets the baseline for further exploration of combining the advantages of event cameras and conventional cameras for multi-animal tracking.

\end{abstract}

\section{Introduction}

Quantification of animal behavior is a critical part of neuro-scientific and biological research. 
The first step towards quantifying animal behavior consists of tracking the animal's movements. 
In recent years a multitude of methods have emerged to leverage recent advances in Computer Vision for visual tracking of animals~\cite{Pereira2020qbb}. However, current tracking systems are limited by the capabilities of the used hardware, like sensors, processors and power supply. Some of these limitations can be overcome by the use of event cameras.

Event cameras \cite{Lichtsteiner08ssc} %
are bio-inspired sensors that differ from conventional frame-based cameras in the way that visual data is acquired. While frame-based cameras capture images at a fixed frame rate, event cameras measure brightness changes at each pixel independently and output them in the form of an event stream (which encodes the pixel location, time and sign of the brightness changes). The different principle of operation endows event cameras with attractive properties over conventional cameras, such as a very high temporal resolution (\si{\micro\second}), a very high dynamic range and low power consumption. We refer to a comprehensive survey for details \cite{Gallego20pami}.

The quantification of animal behavior can be performed in a wide variety of ways ranging from observations in natural conditions to experiments in controlled laboratories. 
Furthermore, there are a multitude of representations of captured movements, which differ in complexity (e.g., centroid tracking vs.~3D animal pose estimation) and number of tracked animals. The wide variety of tracking tasks with their respective modalities places different requirements on the hardware and algorithms. This opens different possibilities for the application of event cameras in animal behavior analysis.

We propose a stereo co-capture system for recording and tracking of animals. Each monocular system consists of a frame-based camera and an event camera, with their views spatially aligned via a beamsplitter.
A hardware trigger provides high precision temporal alignment of all camera signals. 
The system enables a fair comparison with frame-based tracking algorithms and furthermore allows us to perform tracking using both data streams to combine their advantages.

We show an application to fish tracking in a laboratory environment using the stereo co-capture system (\cref{fig:eyecatcher}). 
Acquiring the fast and erratic movement of fish with conventional cameras requires high frame rates. 
The high temporal resolution of event cameras allows us to effectively capture these fast movements. 
In summary, our contributions are:
\begin{itemize}
\item A co-capture system providing temporally and spatially aligned frames and events for animal recording,
\item A baseline multi-object tracking algorithm using events, with an application to fish tracking.
\end{itemize}
\section{Related Work}
\label{sec:relatedwork}

\subsection{Animal Tracking Technology}
\label{sec:related:animal-tracking}

There is a vast variety of methods and tools available for tracking animals. 
They can be roughly categorized according to the movement representation and the number of tracked animals. The simplest form of tracking is the description of an animal movement as a trajectory of points or ellipses, for example obtained by background subtraction or thresholding \cite{Branson2009high}. This method is computationally light and in different variations is widely adopted in many open source tracking tools \cite{Panadeiro2021review}. However, classical methods like this one fail in more difficult scenarios with complex or dynamic backgrounds and occlusions. The technique can be extended to multiple animals, which introduces the problem of identity assignment. 
The general task of multi-object tracking usually follows the tracking-by-detection paradigm. 
In a first step objects are detected in the camera frames; in a second step the detected objects are associated between frames. 
This decouples the tasks of object detection and data association, allowing researchers to adopt state-of-the-art deep-learning--based object detection methods. 
Recently, end-to-end learned approaches, like \cite{Zeng21arxiv} show promising results to further improve tracking accuracy, beyond the tracking-by-detection paradigm.

Generally, the association task can be addressed by modelling the appearance and/or motion of the animals. A simple and widely used approach is described in \cite{Bewley2016sort}. A constant velocity model is assumed, to predict motion using a Kalman filter. Bounding box predictions of the next frame are associated according to their intersection over union (IoU) with the predicted bounding boxes of existing trackers using the Hungarian algorithm. 
The authors of \cite{Naiser18vaib} use an offline method, where tracklets over several frames are built and a learned approach is used for animal re-identification. 
We adopt the algorithm in \cite{Bewley2016sort} and extend it for usage on event data.

\subsection{Event-based Object Tracking}

Event cameras capture pixel-level brightness changes asynchronously, called events.
Assuming constant illumination, events are caused by moving edges \cite{Gallego20pami}. 
This motivates their use for efficient object tracking. 
Early approaches follow a blob-tracking \cite{Litzenberger06itsc} or pattern-tracking \cite{Ni15neco,Lagorce15tnnls} paradigm. These approaches work on a per-event basis, associating incoming events with existing objects, subsequently updating the position according to the associated events. Similar classical approaches are computationally light but mostly tailored to specific applications. The authors of \cite{Afshar20jsen} use a template matching approach to track space objects (satellites, etc.).
A second class of event-based tracking algorithms leverages deep-learning--based approaches. 
In a first step, frame-like representations are obtained from the events, for compatibility with mainstream computer vision methods. Subsequently, methods like CNNs that rely on a grid-like data representation can be used \cite{Iacono18iros}.
The signals obtained from event- and frame-based cameras are complementary, therefore a third class of algorithms proposes to jointly use events and frames to solve the tracking task. The authors of \cite{Gehrig19ijcv} extract features from frames and subsequently tracks them asynchronously using events. In \cite{Liu16iscas} cluster-based event tracking is used to generate regions of interest (ROIs); in a second step a CNN is used to classify the region proposals on the frames.

\subsection{Co-capture Systems}
In the literature we find cameras that jointly capture frames and events, such as the DAVIS \cite{Brandli14ssc,Taverni18tcsii}. 
However, these prototypes have a low spatial resolution and produce low-quality grayscale frames, with a dynamic range of $\approx 55$dB. Recently researchers have resorted to building custom sensing devices, using a beam splitter mirror to spatially align the field of views of an RGB and an event camera \cite{Wang20cvpr,Hidalgo22cvpr}. %

\section{Co-capture System and Fish Tracking Method}
\label{sec:method}

Event cameras offer several interesting properties to overcome limitations of conventional cameras (motion blur, low dynamic range, redundant data in static environments, etc.). 
Under the constant brightness assumption event cameras capture moving edges, 
which are very informative footprints for object tracking tasks.
However, as is usual in an emerging field, there is a lack of datasets and benchmarks, to evaluate the performance of event-based algorithms.

For this reason we propose using a co-capture system to acquire spatially and temporally aligned (e.g., synchronized) data from event and frame-based cameras. In the following, we describe the co-capture system and a basic algorithm for multi-object tracking with event cameras as well as a frame-based baseline algorithm, for comparison.

\subsection{System Specification and Calibration}
\label{sec:method:calibration}
\begin{figure}[t]
\centering
\includegraphics[width=0.85\linewidth]{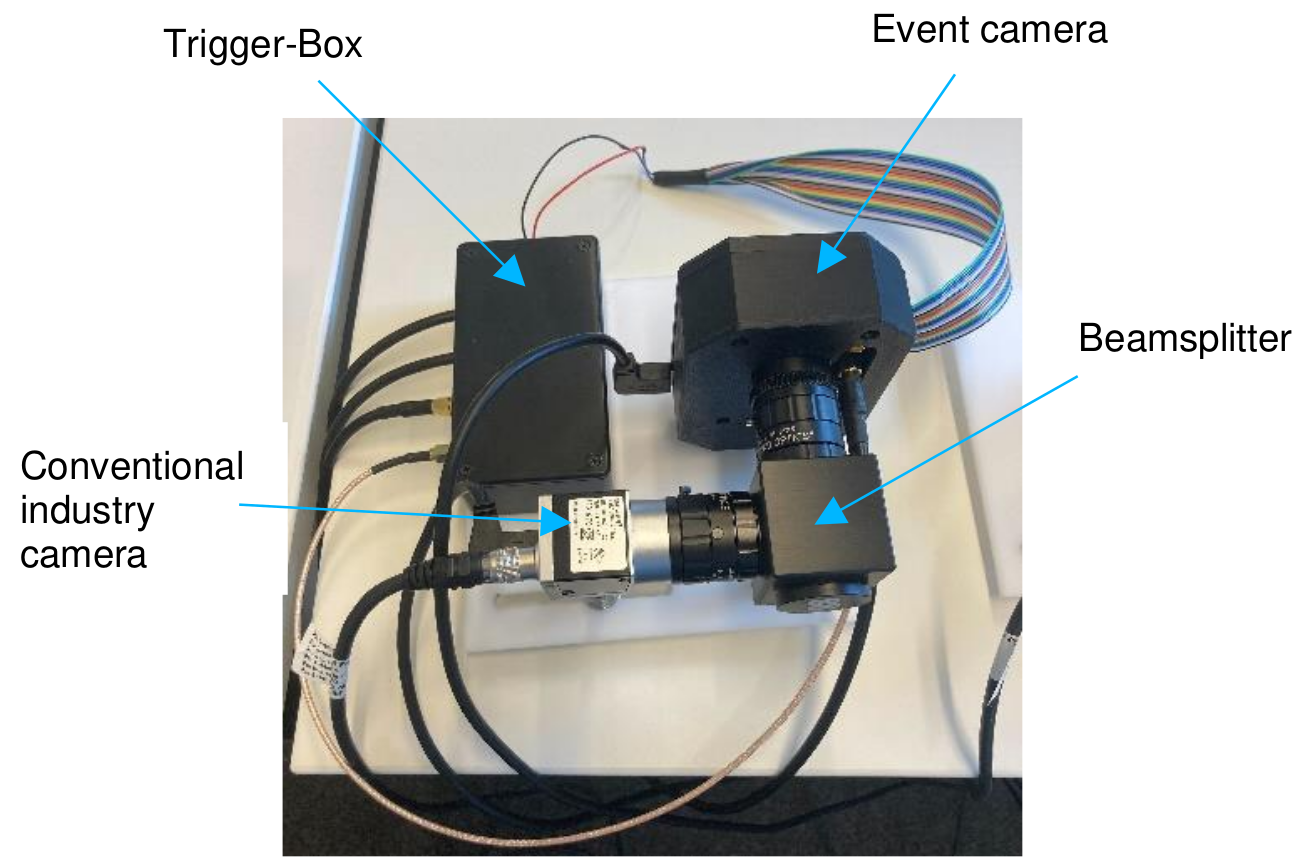}
\caption{Part of our co-capture system. Stereo is shown in \cref{fig:experiments:recording_setup}.}
\label{fig:method:cocapture}
\end{figure}

The co-capture system consists of an event camera (Prophesee EVK3 Gen4.1, 1280 $\times$ 720 pixels), a frame-based camera (Basler acA1300-200um, 1280 $\times$ 1024 pixels), a beamsplitter (Plate Bs C-Mount VIS50R/50T) and a custom-build trigger-box. 
\Cref{fig:method:cocapture} shows the system and its components. 
Every camera receives a synchronized trigger signal from a micro-controller in the trigger-box. 
At each rising edge of the \emph{rect}-signal a grayscale frame is acquired and a timestamp is generated in the event camera. Thereby, the frames can be accurately time-aligned with the event data.

The beamsplitter approximately aligns the field of views of both cameras. 
To achieve a more accurate alignment it is necessary to warp the data from one of the cameras onto the other using the homography between their coordinate systems.
To estimate the planar homography we use a standard calibration software, obtaining the extrinsic and intrinsic calibration of the two cameras (we used \cite{Basalt}). 
Afterwards the homography $H$ can be decomposed as \cite[p.327]{Hartley03book}:
\begin{equation}
    H = R - \frac{\mathbf{t} \mathbf{n}^\top}{d},
\end{equation}
where $R$ is the $3\times3$ rotation matrix, $\mathbf{t}$ is the translation vector between the the two coordinate systems, and $\mathbf{n}$ and $d$ parameterize a world plane of the form $\mathbf{n}^\top\mathbf{x} + d = 0$.
Since the camera centers are very close to each other, we can assume the ratio $\mathbf{t}/d$ to be small and therefore approximate the homography by its rotational part $H \approx R$. The homography is mapping from one pixel-domain to the other.

\subsection{Tracking Method}
\label{sec:method:tracking}

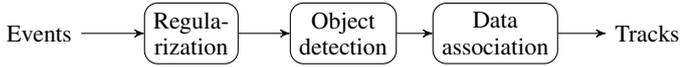
\begin{figure}[t]
\centering
\begin{tikzpicture}[x=1cm, y=1cm, font=\small,>=stealth', scale=0.8]

\node[rounded corners=5pt] (events) at (-2, 0) {Events};
\node[draw, align=center, rounded corners=5pt] (event_frames) at (0.5,0) {Regula-\\rization};
\node[draw, align=center, rounded corners=5pt] (object_detection) at (3,0) {Object\\ detection};
\node[draw, align=center, rounded corners=5pt] (data_association) at (5.5,0) {Data\\ association};
\node[rounded corners=5pt] (tracks) at (8,0) {Tracks};

\draw[->] (events) -- (event_frames);
\draw[->] (event_frames) -- (object_detection);
\draw[->] (object_detection) -- (data_association);
\draw[->] (data_association) -- (tracks);

\end{tikzpicture}
\caption{Processing steps of the event tracking algorithm.}
\label{fig:method:tracking}
\end{figure}

To validate the approach and the comparability of the data, we propose a baseline algorithm to perform event-based multi-animal tracking. 
Our method uses a classical tracking-by-detection approach, combined with common CNN-based object detectors.
The basic pipeline is depicted in \cref{fig:method:tracking}. 

Event-based cameras detect pixel-independent brightness changes. Specifically, they output an event $e_k = (\mathbf{x}_k, t_k, p_k)$ whenever the logarithmic intensity $L(\mathbf{x}, t) = \log I(\mathbf{x}, t)$ changes by a certain threshold. Where $\mathbf{x} = (x, y)^\top$ is the pixel position, $t$ is a timestamp, typically in microsecond resolution and $p_k \in \{-1, +1\}$ signals if the brightness change was positive or negative. 

In the first step of the pipeline (\cref{fig:method:tracking}), we compute time maps $T(x, y)$ from the event stream \cite{Lagorce17pami}, where each pixel in this time map stores the timestamp of the latest event which occurred at that pixel location. 
This step adapts the events into a grid format that is compatible with a large body of algorithms designed for image-based data. Another advantage of this representation is that it can be updated asynchronously on every event and therefore in principle enables processing without loosing temporal accuracy. 
Similarly to \cite{Lagorce17pami}, we apply a pixel-wise exponential decay $\tau$ of the form
\begin{equation}
    I_i(x, y, t_i, p) = e^{-(t_i - T(x, y, p))/\tau}
\end{equation}
where $t_i$ is the current timestamp. 
For each polarity one timestamp map is created, resulting in two output maps, which will be the input channels for the next processing step.
 
After computing the time maps, we apply modern CNN approaches on this representation. 
We use the latest implementation \cite{yolov5} of \cite{Redmon16cvpr}, obtaining $n$ bounding boxes for each frame. 
Subsequent tracking is performed using \cite{Bewley2016sort} (see \ref{sec:related:animal-tracking}).

To compare the methods, the same approach is tested with reconstructed \cite{Rebecq19pami} and grayscale frames, using the same object detector and tracking algorithm. 
For each of the three representations a separate detector is trained using transfer learning with a small hand-labelled dataset of $\approx 150$ snapshots.
\section{Experiments}
\label{sec:experim}

\subsection{Data Acquisition}
\label{sec:experim:dataset}

\begin{figure}[!t]
\centering
\subfloat[]{\includegraphics[height=0.6\linewidth]{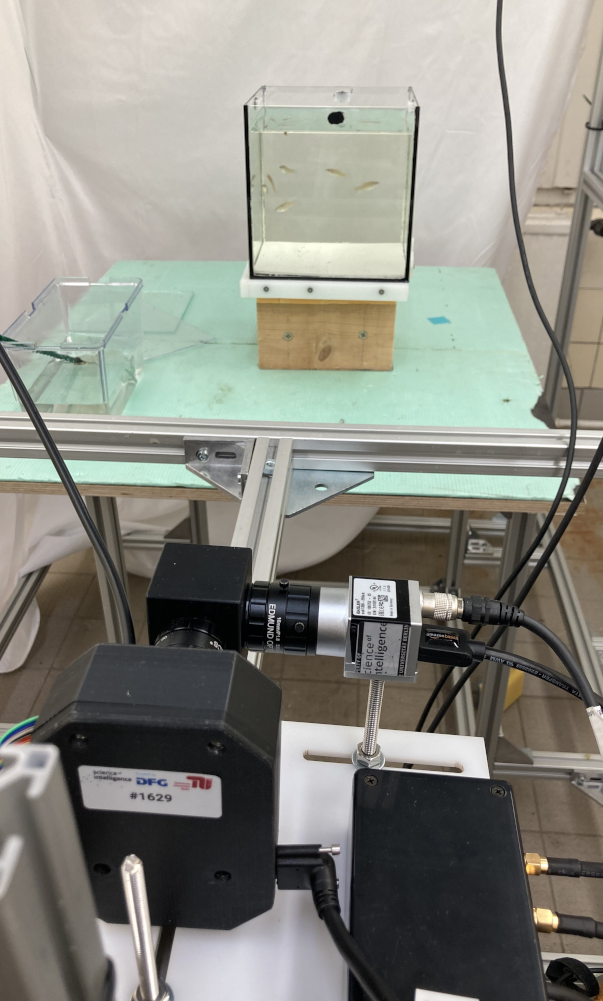}%
\label{fig:experiments:front}}
\hfil
\subfloat[]{\includegraphics[height=0.6\linewidth]{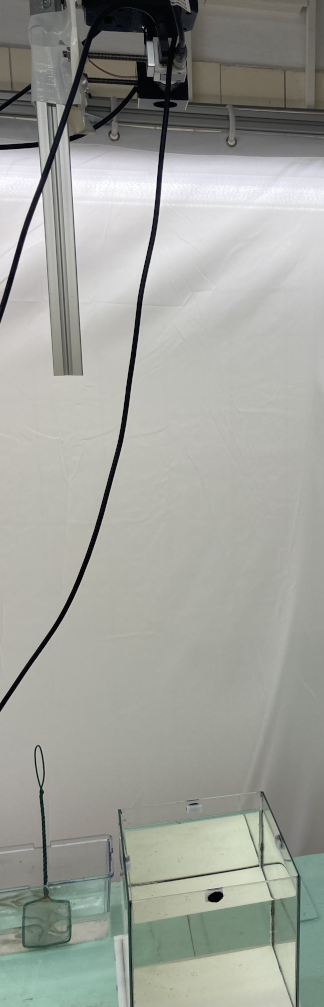}%
\label{fig:experiments:top}}
\caption{The setup for the fish recordings. One co-capture system with front view (a), one with top view (b).}
\label{fig:experiments:recording_setup}
\end{figure}

To show the capabilities of the co-capture system we recorded live fish (\textit{Poecilia formosa}) during ongoing experimental work in the laboratory of Prof. Jens Krause. Two synchronized co-capture systems were used to record fish in a water tank, using one co-capture system from the front and one from the top (see \cref{fig:experiments:recording_setup}). 
The method currently presented uses only the top-view recordings. 
Stereo extensions are planned: tracks from both views will be fused using an EKF to reconstruct the 3D trajectories of the animals.

\begin{figure}[t]
\centering
\gframe{\includegraphics[width=0.5\columnwidth]{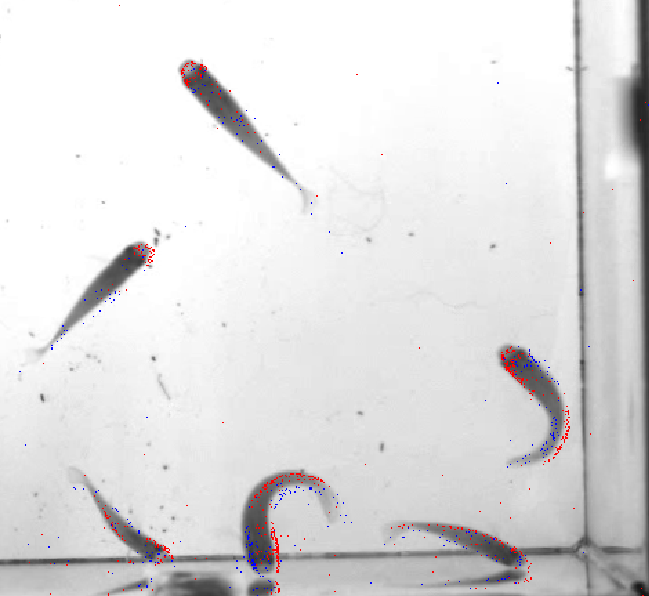}}
\caption{Event-data visualized overlaid on the grayscale frames. 
The blue and red dots represent positive and negative brightness changes (events), respectively.}
\label{fig:experiments:overlay}
\end{figure}

We recorded 12 sequences with 1 to 6 fish. 
The Basler camera recorded at 120 fps. 
The event cameras delivered an average event rate of $675$ thousand events/s. 
The fish were located in a tank of size 20$\times$20$\times$20 cm. 
\Cref{fig:experiments:overlay} shows the overlay of events and grayscale frames.
A visualization of the top view in the three different representations and tracking methods tested can be seen in \cref{fig:eyecatcher}.

\subsection{Results and Discussion}

The goal of the approach is 
($i$) to compare tracking algorithms working on event-based and frame-based data
and ($ii$) to combine events and frames to increase tracking accuracy. 
The asynchronous time maps allow us to increase the frequency at which the trackers are updated up to \si{\micro\second} accuracy. 
The introduced tracking algorithm shows the feasibility of the approach. \Cref{tab:experiments:map} reports the mean average precision of the object detector trained on the different representations and evaluated on a hand-labeled validation dataset. 
The training sets for the grayscale frame and the time-map detector are identical, in the sense that the time maps were queried at the times of the frames and identical annotations were used. 
Furthermore, \cref{tab:experiments:map} presents the average tracklet length of the three approaches validated on three recorded sequences containing 1 to 3 fish.
With more than 10 seconds of average tracking time, the trackers are stable.

\begin{table}[!t]
\renewcommand{\arraystretch}{1.3}
\centering
\begin{tabular}{lccc}
\toprule
                   & gray frames & E2VID & time-surface \\
\midrule
$mAP_{.5:.05:.95}$ & $0.6169$ & $0.4936$ & $0.4224$     \\[0.5ex]
Avg. tracklet time [s] & 20.42  & 16.28 & 14.11        \\
\bottomrule
\end{tabular}
\caption{Mean-average precision and average tracklet time of the object detectors trained on the three different input data.}
\label{tab:experiments:map}
\end{table}

The \emph{mAP} and the tracklet time in the conducted experiment are lower for both event-based representations compared to the gray-scale frames. However, the preliminary results serve as a proof of concept for the chosen approach. They show that event-based multi-animal tracking following the tracking-by-detection paradigm is possible. This sets the baseline for further exploration of event-based tracking. The classical CNN-based object detectors do not exploit the sparse nature of event-data. This motivates the adoption of event-based object detectors for animal tracking.
\label{sec:experim:results}
\section{Conclusion}

We have presented a co-capture system for recording and tracking animals using frames and events.
We have also described a baseline algorithm to use the event data for multi-animal tracking, which provides the base for qualitative comparison and development of advanced tracking algorithms combining the strengths of both sensor types.
With asynchronous object detectors the event data can be used for computationally efficient tracking, 
to capture very fast movements or to perform tracking under challenging lighting conditions. 
We plan to extend and test our method to more challenging scenarios, such as long-term observation of wildlife animals.

\section*{Acknowledgment}
Funded by the Deutsche Forschungsgemeinschaft (DFG, German Research Foundation) under Germany’s Excellence Strategy – EXC 2002/1 “Science of Intelligence” – project number 390523135.
We thank Prof. B. Shi's lab, who helped selecting the beamsplitter and lenses of the system 
and Prof. J. Krause's lab for providing the fish tank setup. All animal care and experimental protocols complied with local and federal laws and guidelines, and were approved by the appropriate governmental authorities (LaGeSo G-0089/21 granted to David Bierbach).

\bibliographystyle{IEEEtran}

\end{document}